\documentclass[sigconf]{acmart}

\AtBeginDocument{%
  }

\usepackage{bbding}
\usepackage{cleveref}
\usepackage{multirow}
\usepackage{booktabs}    
\usepackage{multirow}    
\usepackage{graphicx}    
\usepackage{pifont}      
\usepackage{xcolor}      

\usepackage{amssymb}
\usepackage{pifont}

\usepackage{enumitem}
\definecolor{cmark}{RGB}{34,139,34}

\usepackage[table,xcdraw]{xcolor}
\usepackage{multirow}
\usepackage{tabularx}
\usepackage{float}
\usepackage{graphicx}
\usepackage{amsmath,mathrsfs}
\usepackage{amssymb}
\hypersetup{colorlinks=true, allcolors=red}

\definecolor{best}{RGB}{197,17,17}
\definecolor{second}{RGB}{0,82,180}

\copyrightyear{2026}
\acmYear{2026}
\setcopyright{cc}
\setcctype{by-nc-nd}

\acmConference[MM '26]
{Proceedings of the 34th ACM International Conference on Multimedia}
{November 10--14, 2026}
{Rio de Janeiro, Brazil}

\acmBooktitle{Proceedings of the 34th ACM International Conference on Multimedia
(MM '26), November 10--14, 2026, Rio de Janeiro, Brazil}

\acmISBN{979-8-4007-2213-4/2026/11}
\acmDOI{10.1145/3767308.3836547}

\begin{document}

\title{Stream4D: Streaming 4D Reconstruction with Training-free Motion Decoupling via Causal KV-Cache Attention}

\author{Wei Zhang}
\email{zhangwei707@mail.nwpu.edu.cn}
\orcid{0000-0003-4655-356X}
\affiliation{%
  \department{School of Computer Science}
  \institution{Northwestern Polytechnical University}
  \city{Xi'an}
  \country{China}
}

\author{Ying Feng}
\email{yfeng.von@gmail.com}
\orcid{0009-0006-7838-3103}
\affiliation{%
  \department{vivo BlueImage Lab}
  \institution{vivo Mobile Communication Co., Ltd.}
  \city{Hangzhou}
  \country{China}
}

\author{Mengtong Wu}
\email{wumengtong@mail.nwpu.edu.cn}
\orcid{0009-0004-7797-8864}
\affiliation{%
  \department{School of Computer Science}
  \institution{Northwestern Polytechnical University}
  \city{Xi'an}
  \country{China}
}

\author{Shiqiang Gong}
\email{gongshiqiang@mail.nwpu.edu.cn}
\orcid{0009-0006-9412-9949}
\affiliation{%
  \institution{Northwestern Polytechnical University}
  \city{Xi'an}
  \country{China}
}

\author{Huaqi Zhang}
\email{zhanghuaqi@vivo.com}
\orcid{0000-0001-5891-6204}
\affiliation{%
  \department{vivo BlueImage Lab}
  \institution{vivo Mobile Communication Co., Ltd.}
  \city{Hangzhou}
  \country{China}
}

\author{Qi Zhang}
\authornote{Corresponding author.}
\email{nwpuqzhang@gmail.com}
\orcid{0000-0001-9611-6697}
\affiliation{%
  \department{vivo BlueImage Lab}
  \institution{vivo Mobile Communication Co., Ltd.}
  \city{Hangzhou}
  \country{China}
}

\renewcommand{\shortauthors}{Wei Zhang et al.}


\begin{abstract}
   4D scene reconstruction requires robust handling of dynamic objects, yet existing streaming methods commonly assume static scenes, leading to pose drift and ghosting artifacts over long sequences. We introduce Stream4D, a training-free, motion-decoupled streaming reconstruction framework that can be directly integrated into causal Transformer backbones (e.g., StreamVGGT, Stream3R). Our key observation is that causal attention implicitly encodes rich temporal motion cues within the KV cache. We adaptively aggregate these cues from multiple attention layers and across temporal windows, and fuse them with depth priors to handle challenging cases such as occlusions and textureless regions, producing precise dynamic masks that decouple static and dynamic regions. We then apply a soft key masking scheme that leverages these dynamic masks to suppress interference from dynamic regions in causal attention, thereby improving camera pose estimation accuracy. Stream4D inherits the efficiency of causal attention and remains compatible with existing streaming architectures. On diverse dynamic-scene benchmarks, Stream4D significantly improves camera pose estimation and static-scene reconstruction quality, yields reliable dynamic masks, and effectively mitigates long-sequence ghosting.
\end{abstract}



\begin{teaserfigure}
  \centering
  \includegraphics[width=0.9\textwidth]{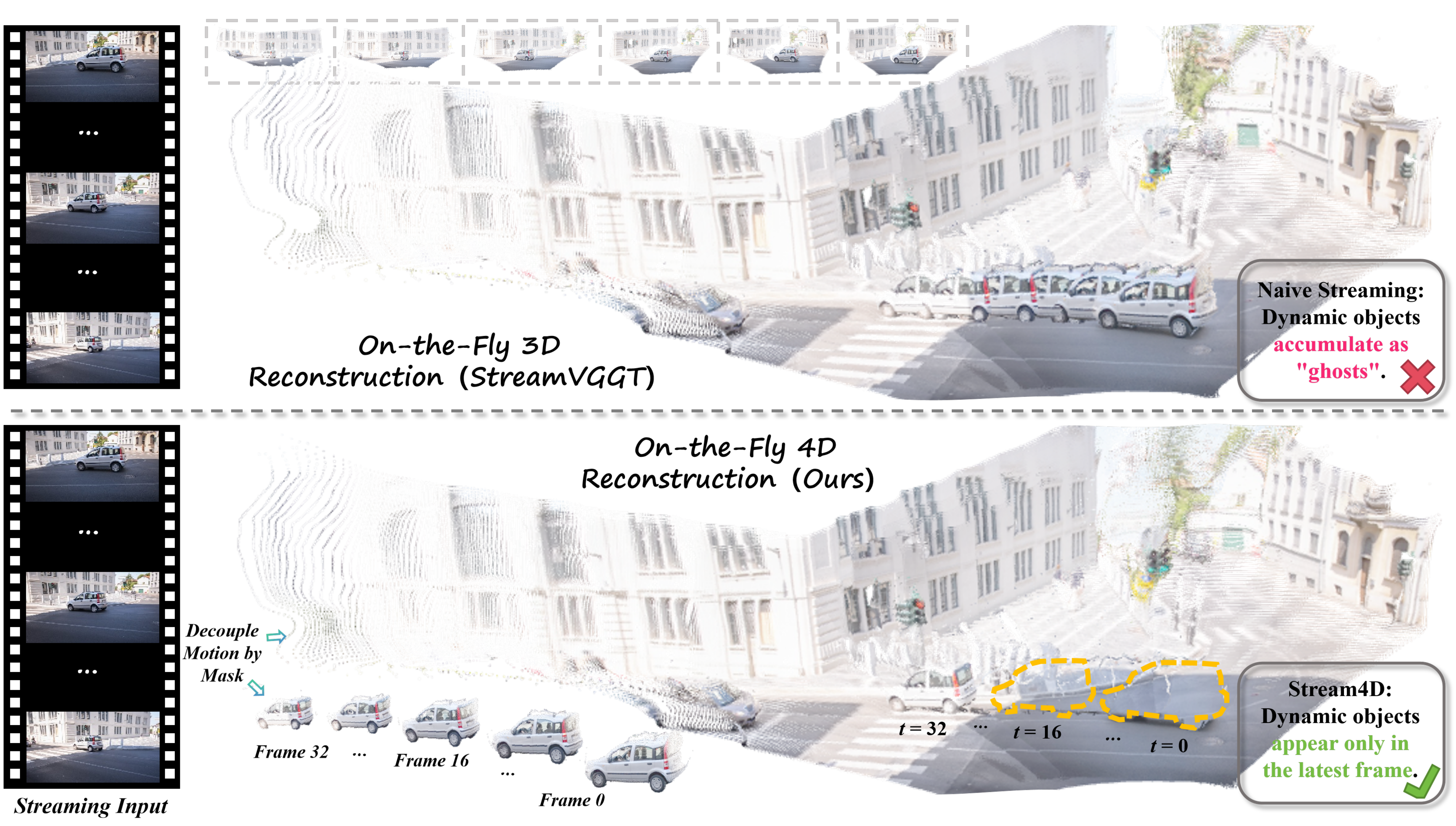}
  \caption{We present Stream4D, a training-free framework that infers dynamic masks from causal-attention cues to decouple motion in streaming reconstruction. By masking dynamic regions during historical aggregation, Stream4D stabilizes online pose estimation and reduces long-sequence ghosting, yielding cleaner dynamic reconstructions.}
  \label{fig:teaser}
\end{teaserfigure}

\maketitle



\section{Introduction}
\label{sec:intro}

4D scene reconstruction is a core capability for embodied intelligence systems~\cite{ozyecsil2017survey, oliensis2000critique, su2026xstreamvggt, zhang2026vggtcd}, with broad applications in autonomous ~\cite{huang2023tri,huang2024sgaussian,zhong2026dyn}, AR/VR~\cite{zheng2024gps,zhong2025adaptive,zhong2025semi}, and robotic navigation~\cite{wang2024embodiedscan,wu2025embodiedocc, wu2025sasw,wu2026promsa}. Traditional SfM/MVS pipelines rely on explicit geometric constraints and iterative optimization~\cite{furukawa2009accurate,gu2020cascade}, making it difficult to balance speed and scalability~\cite{zhang2025semantic, zhang2025refined, 10720843,11595022}. Recent end-to-end learning methods have advanced multi-view 3D prediction, yet pairwise approaches typically require costly global alignment and post-processing, while memory-augmented schemes are prone to error accumulation~\cite{zhang2025review}. Fast3R~\cite{yang2025fast3r} and VGGT~\cite{wang2025vggt} enable dense global interaction via Transformer feed-forward structures, but depend on global self-attention, forcing reprocessing of the entire sequence for each incoming frame. To achieve online and efficient inference, causal architectures such as StreamVGGT~\cite{zhuo2025streaming} and Stream3R~\cite{lan2025stream3r} leverage temporally causal attention and KV-Cache historical memory, substantially reducing computation and storage on long sequences—marking a key step toward scalable 4D vision.

However, existing streaming reconstruction methods commonly assume static scenes and overlook interference from dynamic objects in the real world. Moving objects break photometric consistency and epipolar geometry, mislead camera pose estimation, and accumulate into ghost artifacts over long sequences. Although 3D foundation models like VGGT~\cite{wang2025vggt, wang2024dust3r} can quickly provide high-quality initial estimates, their training and inference typically default to static assumptions and lack motion-decoupling mechanisms. Prior attempts such as VGGT4D~\cite{hu2025vggt4d} and Easi3R~\cite{chen2025easi3r} perform training-free dynamic segmentation via attention statistics, but respectively rely on global bidirectional or pairwise local interactions: the former is incompatible with causal streaming, and shallow-layer attention becomes ineffective under causality constraints; the latter only captures near-frame relations, yielding limited temporal consistency and boundary quality, and fails to prevent dynamic information from accumulating over time. Consequently, stabilizing dynamic-scene reconstruction under causal streaming remains a bottleneck for practical long-sequence 4D reconstruction.

We present \textbf{Stream4D}: a training-free, pluggable, motion decoupled streaming 4D reconstruction framework compatible with causal backbones (e.g., StreamVGGT~\cite{zhuo2025streaming}, Stream3R~\cite{lan2025stream3r}) for dynamic scenes. Our main contributions are as follows:
\begin{itemize}
\item We reveal that causal attention implicitly encodes rich temporal motion cues within the KV cache, and propose an \emph{\textbf{attention-based saliency aggregation}} method that adaptively aggregates these cues from multiple attention layers and across temporal windows.
\item We introduce a robust \emph{\textbf{dynamic mask generation pipeline}} that fuses attention-derived saliency with depth priors, effectively handling challenging cases such as occlusions and textureless regions.
\item We design a \emph{\textbf{soft key masking}} scheme that directly operates on the KV cache to suppress dynamic regions during historical aggregation, improving camera pose estimation accuracy without retraining or architectural modifications.
\item Extensive experiments on dynamic-scene benchmarks demonstrate that \mbox{Stream4D}, without any additional training or fine-tuning, consistently improves pose accuracy, produces reliable dynamic masks, and significantly mitigates long-term ghosting artifacts.
\end{itemize}

\section{Related Work}
\noindent\textbf{Streaming Reconstruction.}
Streaming approaches offer a more scalable alternative for 3D reconstruction problems, such as monocular SLAM pipelines~\cite{wu2025point3r}. Inspired by learned online reconstruction, Spann3R~\cite{wang20243d} extends DUSt3R~\cite{wang2024dust3r} with memory-based updates to achieve incremental reconstruction via recursive memory writes. Fast3R~\cite{yang2025fast3r} and VGGT~\cite{wang2025vggt} replace asymmetric decoders with Transformer-based attention stacks, enabling direct multi-view fusion and delivering strong reconstruction quality. However, these methods rely on global full attention, requiring reprocessing of the entire sequence as length grows, which limits real-time scalability. CUT3R~\cite{wang2025continuous} adopts an RNN-style architecture to incrementally process unstructured inputs, but suffers from limited memory capacity and poor compatibility with modern hardware acceleration. Recently, StreamVGGT~\cite{zhuo2025streaming} and Stream3R~\cite{lan2025stream3r} re-conceptualize point-graph prediction as a decoder-only Transformer~\cite{vaswani2017attention,xiao2026staying} task, achieving efficient causal inference via techniques like KV-Cache and windowed attention, scaling well to long sequences while remaining compatible with modern LLM-style training infrastructure. Nonetheless, these streaming methods generally assume static scenes and lack mechanisms to handle dynamic objects, constraining applicability in real-world scenarios~\cite{COMBINER,ReTrack,xiao2025visual,xiao2026not,xiao2026layer}.

\noindent\textbf{4D Scene Reconstruction.}
Early 4D reconstruction methods typically rely on RGB-D sensors~\cite{newcombe2011kinectfusion, li2026ragtrack}, limiting deployment in real-world settings~\cite{xu2026psgait,wu2026language,wu2025dagait,zhong2026semi,huang2026detecting}. Many approaches jointly estimate depth, pose, and residual motion, often using motion masks on static regions, including self-supervised variants and test-time optimization schemes such as CasualSAM~\cite{zhang2022structure} and Robust-CVD. More recent pipelines, e.g., MegaSaM~\cite{li2025megasam}, leverage strong monocular depth priors~\cite{yang2024depth} to achieve robust pose estimation and reconstruction on dynamic videos, while Uni4D~\cite{yao2025uni4d} combines multiple vision foundation models with multi-stage bundle adjustment for high-quality 4D results. Although effective, these pipelines depend on powerful priors and extensive test-time optimization, which hinders scalability. Recent work reduces or replaces test-time optimization using learned priors or attention mechanisms: MonST3R~\cite{zhang2024monst3r} fine-tunes on dynamic data and exploits optical flow; DAS3R~\cite{xu2024das3r} augments the DPT head to produce feed-forward motion masks; CUT3R ~\cite{wang2025continuous} fine-tunes MASt3R~\cite{leroy2024grounding} on mixed static/dynamic data for fast reconstruction; Easi3R~\cite{chen2025easi3r} adaptively modulates attention at inference to perform training-free 4D reconstruction on DUSt3R~\cite{wang2024dust3r}. Despite these advances, most pipelines still require fine-tuning or second-stage post-processing, and training-free variants are often tightly coupled to specific backbones. More critically, these methods are designed around global bidirectional Transformer architectures, making them incompatible with causal streaming and leaving the long-sequence accumulation of dynamic information unresolved. In contrast, our Stream4D is the first training-free 4D reconstruction framework tailored for causal streaming architectures. It achieves motion decoupling via selective KV-Cache updates, supports online reconstruction over long sequences, and avoids ghost accumulation.

\section{Method}

Given a casually captured video sequence $\{ \mathbf{I}_t \in \mathbb{R}^{H \times W \times 3} \}_{t=1}^T$, our goal is to estimate dynamic-object masks, camera motion, and a static scene point cloud. The dynamic-object masks are represented as a segmentation sequence $\{ \mathbf{M}_t \}_{t=1}^T$. The camera motion is represented as the sequence of extrinsic and intrinsic parameters $\{ (\mathbf{P}_t, \mathbf{K}_t) \}_{t=1}^T$. The static scene point cloud is denoted by $\mathcal{X}^\ast$. We first introduce how a causal streaming architecture enables efficient incremental processing via KV Cache (Sec.~\ref{sec:streaming}). Next, we show how to aggregate motion cues from attention layers and generate dynamic masks (Sec.~\ref{sec:Segmentation}). Finally, we present motion-decoupled 4D reconstruction using the dynamic masks (Sec.~\ref{sec:Reconstruction}).

\subsection{Streaming 4D Geometry Reconstruction}
\label{sec:streaming}

Given an image sequence $\mathbf{I}_t$, a generic 4D reconstruction pipeline can be written as
\begin{equation}
\mathbf{F}_t = \mathrm{Encoder}(\mathbf{I}_t), \quad
\mathbf{G}_t = \mathrm{Decoder}(\mathbf{F}_t), \quad
(\mathbf{P}_t, \mathbf{C}_t) = \mathrm{Head}(\mathbf{G}_t),
\end{equation}
where the encoder maps the input frame to image tokens $\mathbf{F}_t \in \mathbb{R}^{N \times C}$, the decoder fuses multi-view information to produce geometry tokens $\mathbf{G}_t \in \mathbb{R}^{N \times C}$, and the prediction head outputs a point map $\mathbf{P}_t \in \mathbb{R}^{3 \times H \times W}$ and a confidence map $\mathbf{C}_t \in \mathbb{R}^{H \times W}$. Existing methods differ in how they aggregate multi-view information in the decoder: pairwise methods (e.g., DUSt3R~\cite{wang2024dust3r}, MASt3R~\cite{leroy2024grounding}) process image pairs via cross-attention; memory-augmented methods (e.g., Spann3R~\cite{wang20243d}, CUT3R~\cite{wang2025continuous}) maintain an external memory for incremental updates; and global interaction methods (e.g., Fast3R~\cite{yang2025fast3r}, VGGT~\cite{wang2025vggt}) adopt all-to-all self-attention to achieve the highest accuracy but require reprocessing the entire sequence for each incoming frame, thus cannot support online streaming inference.

\begin{figure}[t]
    \centering
    \includegraphics[width=1.0\linewidth]{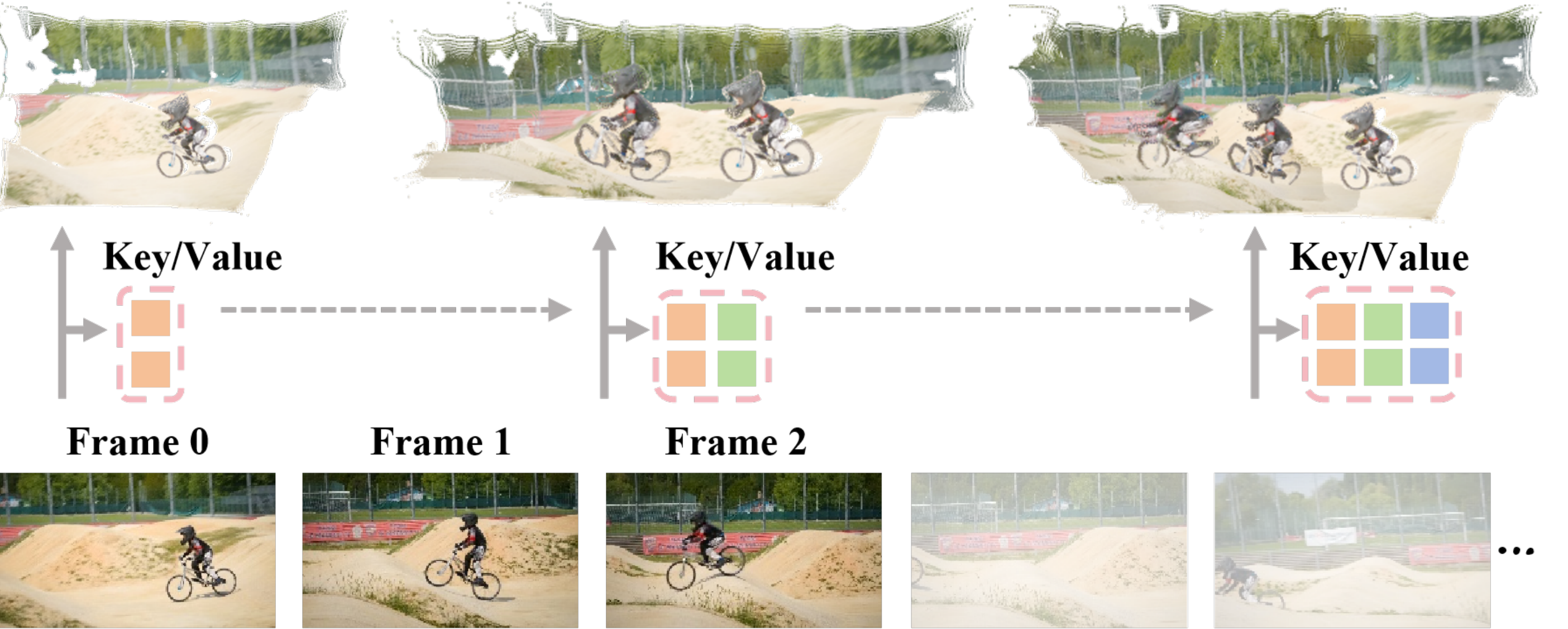}
    \caption{Streaming reconstruction overview. Each incoming frame is encoded into tokens and decoded with temporal causal attention that attends only to past frames via a KV Cache. The cache is incrementally updated with $(\mathbf{K}_t, \mathbf{V}_t)$, enabling efficient online inference.}
    \label{fig:streaming}
\end{figure}

\noindent\textbf{Causal streaming architecture.}
To enable efficient streaming reconstruction, recent works (e.g., StreamVGGT~\cite{zhuo2025streaming}, Stream3R~\cite{lan2025stream3r}, InfiniteVGGT~\cite{yuan2026infinitevggt}) introduce causal Transformer architectures, where temporal causal attention restricts each frame to attend only to past frames:
\begin{equation}
\{ \mathbf{G}_t \}_{t=1}^T = \mathrm{Decoder}\!\left(\mathrm{Causal\text{-}Attn}\left(\{ \mathbf{F}_t \}_{t=1}^T\right)\right),
\end{equation}
and utilize a KV Cache to store historical key-value pairs
$\mathcal{C}_t = \{ (\mathbf{K}_i, \mathbf{V}_i) \}_{i=1}^{t-1}$,
avoiding redundant computation. At inference time, the model accesses cached history via cross-attention:
\begin{equation}
\mathbf{G}_t = \mathrm{Decoder}\!\left(\mathrm{CrossAttn}(\mathbf{F}_t, \mathcal{C}_{t-1})\right), \quad
\mathcal{C}_t = \mathcal{C}_{t-1} \cup \{ (\mathbf{K}_t, \mathbf{V}_t) \}.
\end{equation}

This design enables online incremental reconstruction (Fig.~\ref{fig:streaming}). However, existing causal architectures generally assume static scenes and lack mechanisms to handle dynamic objects. Dynamic objects can mislead pose estimation and geometric modeling and, critically, accumulate in the KV Cache over long sequences, producing ghosting artifacts.

\subsection{Streaming Dynamic Segmentation}
\label{sec:Segmentation}

\begin{figure*}[t]
    \centering
    \includegraphics[width=1.0\linewidth]{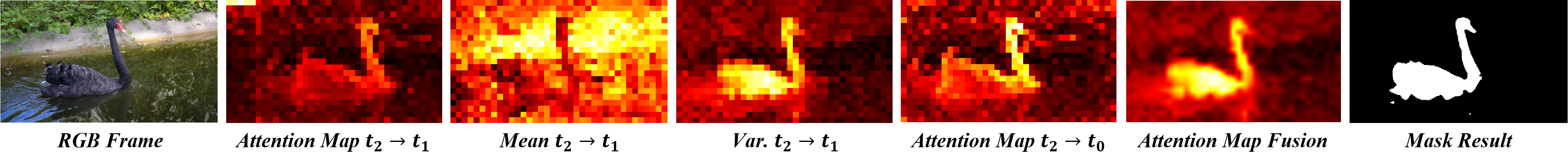}
    \caption{Temporal saliency aggregation and fusion visualization. We compare attention statistics across frames and layers: per-token mean and variance heatmaps revealing low-mean/high-variance patterns on dynamic objects; cross-frame attention maps highlighting temporal inconsistency; multi-scale attention fusion; and final fused saliency and binary masks.}
    \label{fig:heatmap}
\end{figure*}

\begin{figure*}[t]
    \centering
    \includegraphics[width=1.0\linewidth]{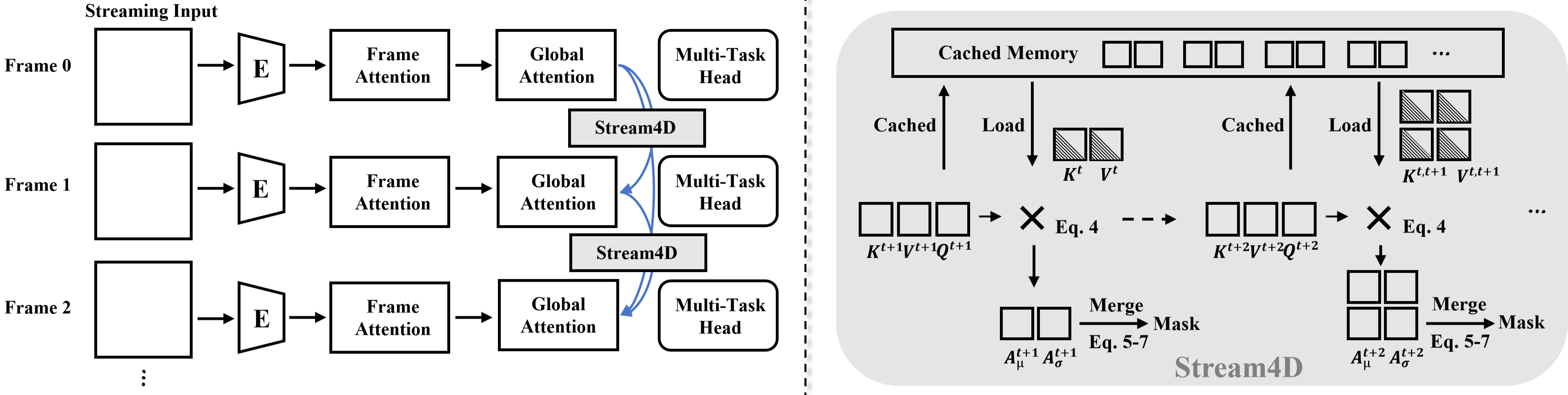}
    \caption{Streaming dynamic segmentation architecture. Left: standard causal streaming reconstruction with a KV Cache, where each frame attends only to past tokens. Right: our Stream4D augments the causal pipeline with motion-decoupled masking. Queries load cached keys/values to compute attention scores, aggregate temporal saliency across frames, and produce dynamic masks. These masks are then applied to modify the cached keys, effectively suppressing interference from dynamic regions during historical aggregation.}
    \label{fig:architecture}
\end{figure*}

We propose a training-free dynamic segmentation method tailored to causal streaming architectures, which aggregates temporal motion cues directly from the KV Cache. Unlike global bidirectional methods (e.g., VGGT4D) or pairwise local methods (e.g., Easi3R) limited to adjacent frames, our approach achieves global temporal consistency via adaptive historical-frame selection and multi-modal fusion with depth priors (Fig.~\ref{fig:architecture}).

\noindent\textbf{Attention-based saliency aggregation.}
In streaming inference, current-frame queries $\mathbf{Q}_t$ attend to cached historical keys via:

$\mathbf{A}_t^{(l)} = \mathrm{Softmax}(\mathbf{Q}_t^{(l)} [\mathbf{K}_1^{(l)}, \ldots, \mathbf{K}_{t-1}^{(l)}]^\top / \sqrt{d})$, where $l$ indexes the layer. Dynamic objects exhibit high-variance, low-mean attention patterns due to spatial motion, while static regions show low variance and high mean (Fig.~\ref{fig:heatmap}). We exploit this by computing per-token statistics $\mu_{t,i}^{(l)}, \sigma_{t,i}^{(l)}$ over an adaptive historical set $\mathcal{H}_t \subseteq \{1, \ldots, t-1\}$, aggregating across frames and layers to obtain dynamic saliency:
\begin{equation}
\mathbf{S}_t^{(l)} = (1 - \bar{\mu}_t^{(l)}) \odot \bar{\sigma}_t^{(l)}, \quad
\mathbf{S}_t^{\mathrm{attn}} = \sum_{l \in \mathcal{L}} w_l \cdot \mathbf{S}_t^{(l)},
\end{equation}
where $\mathcal{L}$ spans shallow, mid-level, and deep layers with weights $w_l$ favoring mid-level semantics.

\noindent\textbf{Depth-based complementary cues.}
To handle challenging cases such as occlusions and textureless regions, we fuse attention saliency with depth priors from predicted depth maps $\mathbf{D}_t$. A foreground prior assumes moving objects are closer:
\begin{equation}
\mathbf{S}_t^{\mathrm{depth\text{-}prior}} = \mathrm{Sigmoid}(\beta \cdot (\mathrm{Quantile}(\mathbf{D}_t, q) - \mathbf{D}_t)),
\end{equation}
and a temporal change measure captures inconsistency over a window $\tau$:
\begin{equation}
\mathbf{S}_t^{\mathrm{depth\text{-}change}} = \mathrm{Normalize}\!\left(\frac{1}{\tau} \sum_{i=t-\tau}^{t-1} |\mathbf{D}_t - \mathbf{D}_i|\right).
\end{equation}
The final saliency fuses attention and depth scores:
\begin{equation}
\mathbf{S}_t = \alpha_{\mathrm{attn}} \cdot \mathbf{S}_t^{\mathrm{attn}} + \alpha_{\mathrm{depth}} \cdot (\alpha_{\mathrm{prior}} \cdot \mathbf{S}_t^{\mathrm{depth\text{-}prior}} + \alpha_{\mathrm{change}} \cdot \mathbf{S}_t^{\mathrm{depth\text{-}change}}),
\end{equation}
where fusion weights $\alpha_{\mathrm{attn}}, \alpha_{\mathrm{depth}}, \alpha_{\mathrm{prior}}, \alpha_{\mathrm{change}}$ are empirically determined. The fused saliency is then processed by temporal smoothing $\tilde{\mathbf{S}}_t = \frac{1}{\tau+1} \sum_{k=t-\tau}^{t} \mathbf{S}_k$, Otsu thresholding~\cite{yang2020improved}, and morphological refinement to produce the binary mask $\mathbf{M}_t$.

\subsection{Motion-Decoupled Streaming Reconstruction}
\label{sec:Reconstruction}

With dynamic masks $\{\mathbf{M}_t\}$ from Sec.~\ref{sec:Segmentation}, we decouple static reconstruction from dynamic interference. The core mechanism is \textbf{soft key masking}, which directly operates on the KV Cache to suppress dynamic regions during historical aggregation.

\noindent\textbf{Soft key masking.}
We directly modify the cached keys in the KV Cache by applying a soft suppression based on the dynamic masks:
\begin{equation}
\mathcal{C}_t^{\mathrm{masked}} = \{ (\tilde{\mathbf{K}}_i, \mathbf{V}_i) \}_{i=1}^{t-1}, \quad \text{where} \quad \tilde{\mathbf{K}}_i = \mathbf{K}_i \odot (\mathbf{1} - \gamma \cdot \mathbf{M}_i),
\end{equation}
where $\mathbf{M}_i \in \{0,1\}^{N}$ marks dynamic tokens and $\gamma \in (0,1)$ controls the suppression strength. The decoder then performs temporal attention using the modified cache:
\begin{equation}
\mathbf{G}_t = \mathrm{Decoder}\!\left(\mathrm{CrossAttn}(\mathbf{F}_t, \mathcal{C}_{t-1}^{\mathrm{masked}})\right).
\end{equation}
By directly suppressing dynamic keys in the cache, this scheme prevents dynamic features from interfering with historical aggregation, thereby improving camera pose estimation accuracy. We apply soft key masking to the first few decoder layers where low-level geometric features dominate, and maintain a mask cache to ensure consistent suppression across the sequence.

\noindent\textbf{Static-region geometric constraints.}
Following standard practice, we restrict photometric and geometric consistency losses to static regions identified by the mask:
\begin{equation}
\mathcal{L}(t) = \sum_{i \in \mathcal{H}_t} \mathbf{C}_t \odot (\mathbf{1} - \mathbf{M}_t) \odot \left( \lambda_{\mathrm{photo}} \cdot \rho(\mathbf{I}_t - \mathbf{I}_i^\pi) + \lambda_{\mathrm{geom}} \cdot \|\mathbf{P}_t - \mathbf{P}_i^\pi\|_1 \right),
\end{equation}
where $\mathbf{I}_i^\pi, \mathbf{P}_i^\pi$ denote warped photometric and geometric predictions, and $\mathbf{C}_t$ is the confidence map. The static point cloud $\mathcal{X}^\ast$ is updated incrementally by inserting high-confidence static points after outlier filtering.

\begin{table*}[t]
    \centering
    \caption{Dynamic object segmentation on DAVIS benchmarks. We report mean IoU ($\mathcal{J}_M$) and IoU recall ($\mathcal{J}_R$). ``Type'' indicates Offline vs.\ Streaming (single-pass, causal). $^*$ indicates our adapted implementation with segmentation capability, as the original methods do not support segmentation.}
    \label{tab:davis_seg}
    \resizebox{0.9\linewidth}{!}{
    \scriptsize
        \begin{tabular}{lccccccc}
            \toprule
            \multirow{2}{*}{Method} & \multirow{2}{*}{Type} & \multicolumn{2}{c}{DAVIS-2016} & \multicolumn{2}{c}{DAVIS-2017} & \multicolumn{2}{c}{DAVIS-all} \\
            \cmidrule(lr){3-4} \cmidrule(lr){5-6} \cmidrule(lr){7-8}
            & & $\mathcal{J}_M$ & $\mathcal{J}_R$ & $\mathcal{J}_M$ & $\mathcal{J}_R$ & $\mathcal{J}_M$ & $\mathcal{J}_R$ \\
            \midrule
            DUSt3R~\cite{wang2024dust3r}            & Offline & 42.10 & 45.70 & 35.20 & 35.30 & 35.90 & 34.00 \\
            Easi3R\_dust3r~\cite{chen2025easi3r}    & Offline & 50.10 & 55.77 & 46.86 & 50.54 & 44.10 & 50.85 \\
            Easi3R\_monst3r~\cite{chen2025easi3r}   & Offline & 54.93 & 68.00 & 54.75 & 66.16 & 51.64 & 63.06 \\
            MonST3R~\cite{zhang2024monst3r}          & Offline & 40.42 & 40.39 & 38.07 & 36.05 & 36.98 & 34.52 \\
            DAS3R~\cite{xu2024das3r}             & Offline & 41.13 & 38.67 & 44.51 & 43.95 & 43.33 & 38.93 \\
            VGGT4D~\cite{hu2025vggt4d}           & Offline & 62.12 & 76.80 & 56.45 & 65.62 & 50.75 & 55.59 \\
            \midrule
            CUT3R$^*$~\cite{wang2025continuous} & Streaming  & 28.45 & 18.76 & 31.28 & 27.92 & 27.65 & 23.18 \\
            Stream3R$^*$~\cite{lan2025stream3r} & Streaming  & 32.18 & 21.34 & 35.07 & 31.45 & 30.82 & 27.06 \\
            StreamVGGT$^*$~\cite{zhuo2025streaming} & Streaming  & 34.91 & 23.52 & 37.33 & 33.86 & 33.20 & 29.24 \\
            \textbf{Stream4D (ours)}       & Streaming  & \textbf{43.39} & \textbf{39.34} & \textbf{46.45} & \textbf{42.90} & \textbf{43.06} & \textbf{41.32} \\
            \bottomrule
          \end{tabular}
    }
\end{table*}

\begin{figure*}[t]
    \centering
    \includegraphics[width=1.0\linewidth]{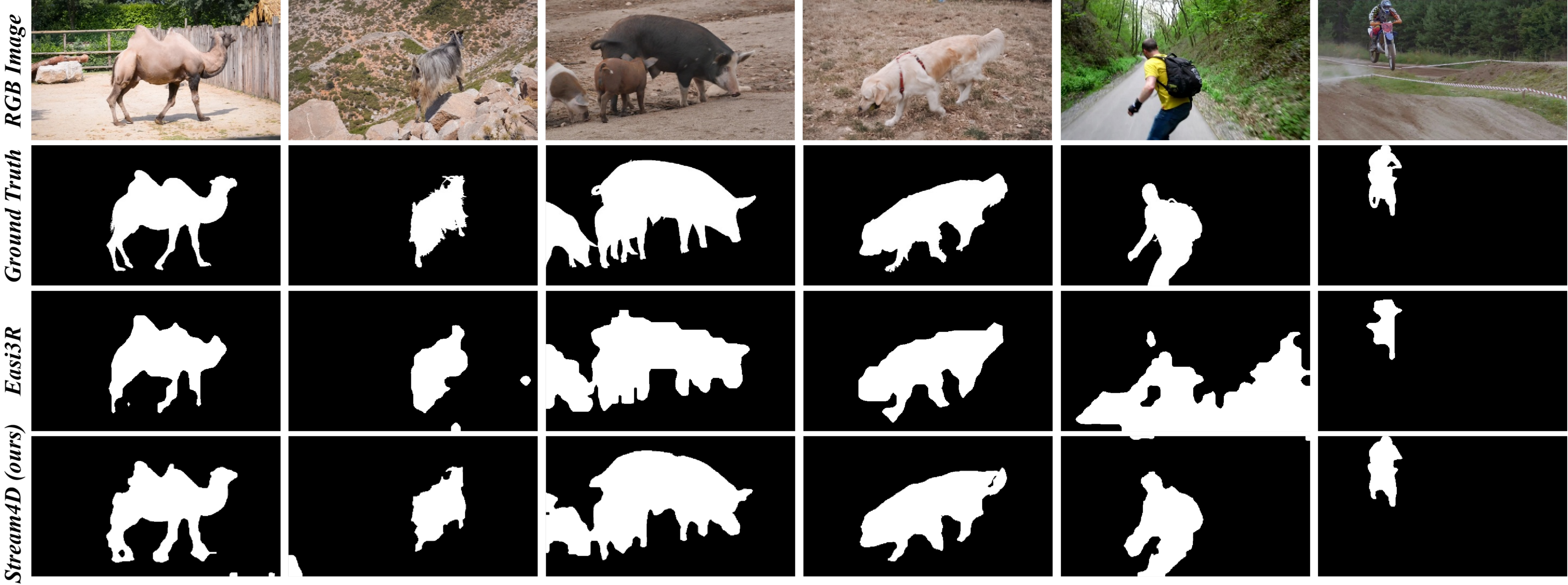}
    \caption{Qualitative results on DAVIS dataset. Each panel shows the input frame, the Easi3R mask, and our Stream4D mask. Easi3R exhibits temporal jitter and boundary breaks under occlusions and non-rigid motions. Our streaming multi-level temporal aggregation with consistency constraints produces temporally smooth and spatially complete masks across slow motions (camel, goat, etc.) and fast motions (skateboarder, motorcyclist), demonstrating robust segmentation under causal KV-Cache attention.}
    \label{fig:qual_seg}
\end{figure*}

\begin{table*}[t]
    \centering
    \small
    \caption{4D reconstruction on DyCheck. We report pose metrics (ATE, RTE, RRE); lower is better. ``Type'' indicates Offline vs.\ Streaming (single-pass, causal).}
    \label{tab:dycheck_recon}
    \footnotesize
    \resizebox{0.8\linewidth}{!}{
    \scriptsize
    \begin{tabular}{lccccc}
        \toprule
        Method & Type & ATE$\downarrow$ & RTE$\downarrow$ & RRE$\downarrow$ \\
        \midrule
        Easi3R\_monst3r~\cite{chen2025easi3r}     & Offline   & 0.032 & 0.008 & 1.075 \\
        Easi3R\_dust3r~\cite{chen2025easi3r}       & Offline   & 0.022 & 0.009 & 0.806 \\
        MonST3R~\cite{zhang2024monst3r}             & Offline   & 0.038 & 0.010 & 1.172 \\
        POMATO~\cite{zhang2025pomato}              & Offline   & 0.128 & 0.027 & 3.648 \\
        SpatialTrackerV2~\cite{xiao2025spatialtrackerv2}    & Offline   & 0.011 & 0.006 & 0.347 \\
        DAS3R~\cite{xu2024das3r}               & Offline   & 0.052 & 0.012 & 1.560 \\
        VGGT~\cite{wang2025vggt}                & Offline   & 0.013 & 0.008 & 0.418 \\
        VGGT4D~\cite{hu2025vggt4d}              & Offline   & 0.010 & 0.007 & 0.374 \\
        \midrule
        CUT3R~\cite{wang2025continuous}               & Streaming & 0.036 & \textbf{0.013} & 0.860 \\
        StreamVGGT~\cite{zhuo2025streaming}           & Streaming & 0.049 & 0.032 & 4.127 \\
        Stream3R~\cite{lan2025stream3r}            & Streaming & 0.015 & 0.018 & 0.545 \\
        Stream4D (ours)     & Streaming &
          \textbf{0.014} \scriptsize{(\textcolor{green}{$\downarrow$6.7\%})} &
          {0.017} \scriptsize{(\textcolor{green}{$\downarrow$5.6\%})} &
          \textbf{0.539} \scriptsize{(\textcolor{green}{$\downarrow$1.1\%})} \\
        \bottomrule
      \end{tabular}
    }
\end{table*}

\begin{figure*}[t]
    \centering
    \includegraphics[width=1.0\linewidth]{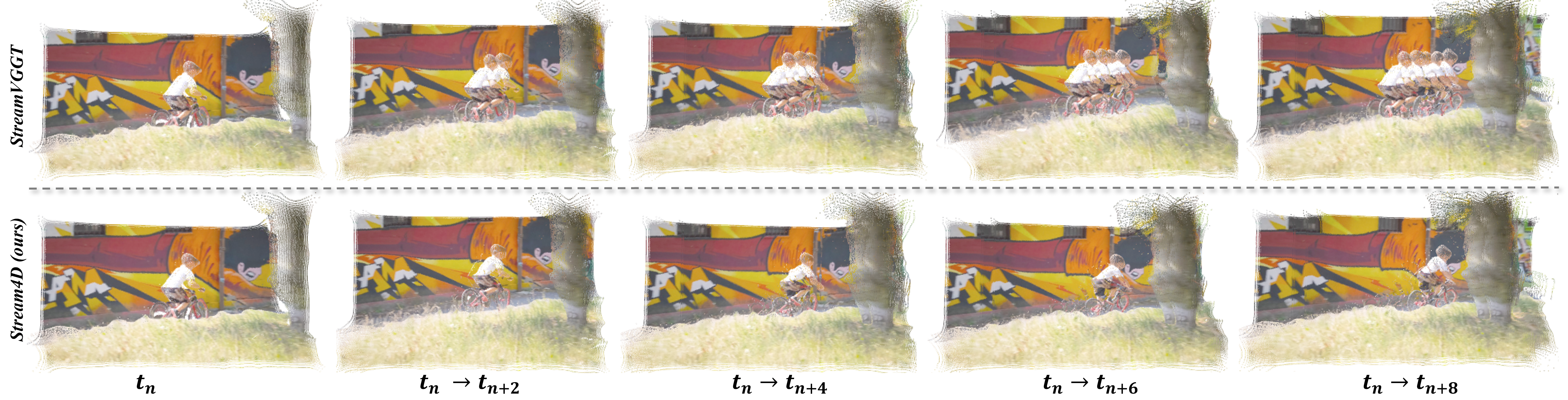}
    \caption{Ghosting suppression in streaming 4D reconstruction. 
    Naïve causal streaming methods integrate geometry from all past frames, causing moving objects to be repeatedly fused into the scene and leaving stacked replicas along their trajectories (ghosting artifacts), which grow with sequence length and can eventually corrupt nearby static surfaces. 
    In contrast, Stream4D derives training-free dynamic masks from causal attention cues and applies soft key masking to suppress dynamic regions in the KV cache, so that dynamic objects appear only in the current frame while the background remains clean and sharp even in long sequences.}
    \label{fig:ghosting}
\end{figure*}

\section{Experiments}

We evaluate Stream4D across three core tasks: dynamic object segmentation (Sec.~\ref{Dynamic}), 4D reconstruction (Sec.~\ref{4D}), and long-sequence streaming efficiency (Sec.~\ref{Efficiency}), followed by ablations. Our protocol emphasizes real-time settings, causal inference, and training-free integration atop pretrained streaming backbones.

\noindent\textbf{Baselines.}
We compare against state-of-the-art streaming reconstruction methods and training-free 4D approaches. StreamVGGT~\cite{zhuo2025streaming} and Stream3R~\cite{lan2025stream3r} are representative causal streaming architectures but assume static scenes; VGGT4D~\cite{hu2025vggt4d} and Easi3R~\cite{chen2025easi3r} perform training-free dynamic segmentation under global bidirectional designs. Stream4D applies training-free motion decoupling directly on pretrained causal streaming models, requiring no fine-tuning or additional data.

 
\noindent\textbf{Datasets.}
We assess dynamic object segmentation on the DAVIS-2016~\cite{perazzi2016benchmark}, DAVIS-2017~\cite{pont20172017}, and DAVIS-all benchmarks, following the official protocols. For 4D reconstruction and camera pose estimation, we evaluate on the DyCheck dataset~\cite{gao2022monocular}, reporting the pose error metrics (ATE, RTE, RRE). Additionally, to validate scalability and efficiency, we conduct performance benchmarks on sequences with varying lengths ranging from 10 to 200 frames.
\begin{figure*}[t]
    \centering
    \includegraphics[width=0.76\linewidth]{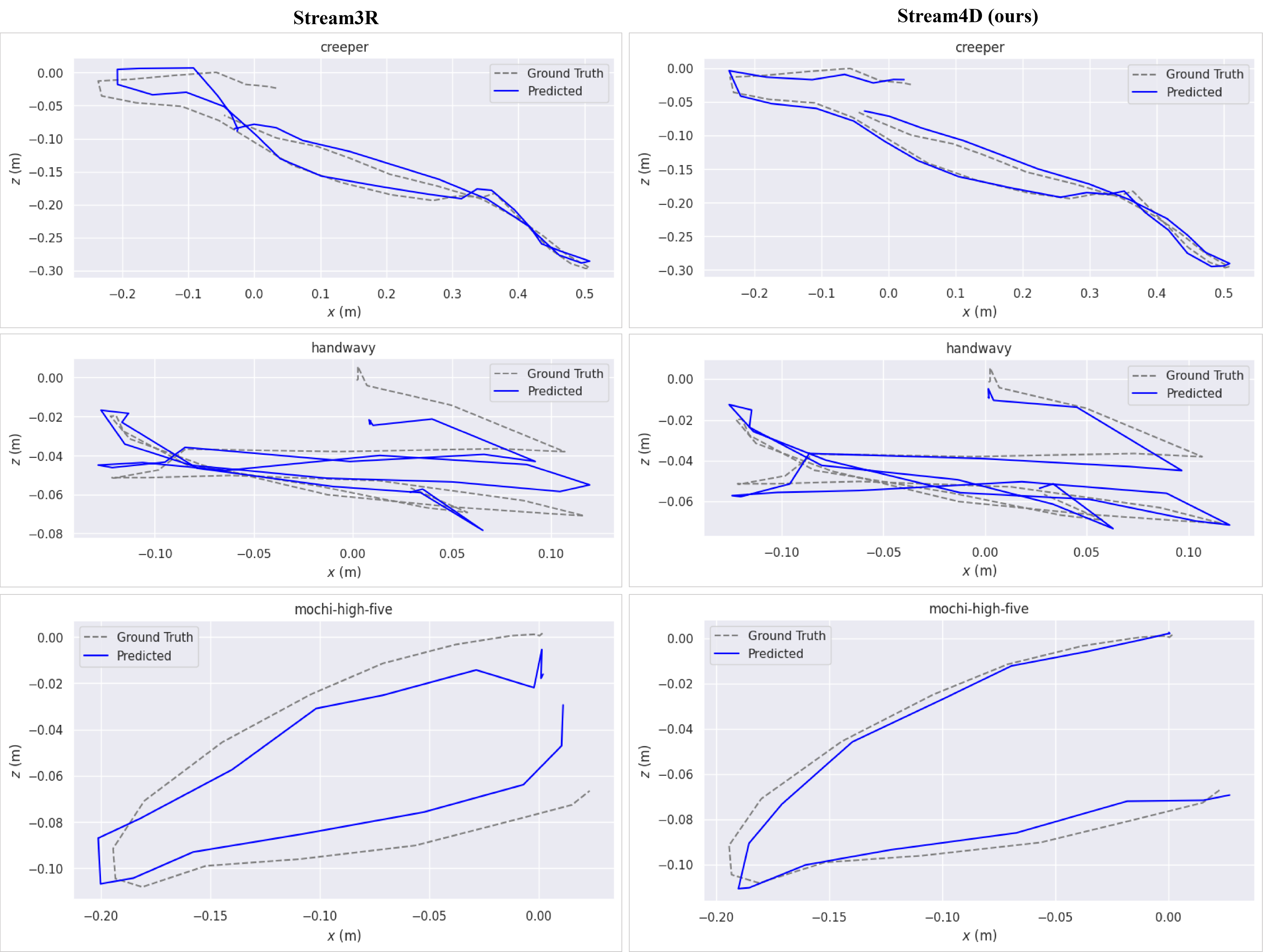}
    \caption{Qualitative comparison of estimated camera trajectories on three DyCheck sequences (\textit{creeper}, \textit{handwavy}, \textit{mochi-high-five}). We plot the predicted trajectory (solid blue) against the ground truth (dashed gray) in the $x$-$z$ plane. Left: Stream3R exhibits noticeable drift and erratic deviations, particularly in sequences with large dynamic motion (\textit{handwavy}, \textit{mochi-high-five}), where moving objects corrupt the pose estimation. Right: our Stream4D produces trajectories that more closely follow the ground truth, demonstrating that motion-decoupled masking effectively mitigates the interference of dynamic regions on camera pose estimation.}
    \label{fig:qual_recon_comp}
\end{figure*}

\subsection{Dynamic Object Segmentation}
\label{Dynamic}

We evaluate dynamic object segmentation on DAVIS-2016~\cite{perazzi2016benchmark}, DAVIS-2017~\cite{pont20172017}, and DAVIS-all following the official DAVIS protocol and splits. Metrics are the mean IoU $\mathcal{J}_M$ and the IoU recall $\mathcal{J}_R$, which jointly measure segmentation accuracy and robustness. All online methods are strictly single-pass and forbid future-frame access to comply with causal streaming constraints. Our method and the causal baselines use identical resolution and preprocessing to ensure a controlled comparison. Since the original StreamVGGT~\cite{zhuo2025streaming} lacks a segmentation head, we integrate the temporal attention aggregation to produce saliency maps, then apply temporal consistency filtering and thresholding to generate binary masks. Easi3R\_online keeps the original implementation and parameters, with the only change of disabling historical-frame access to match causal streaming. Offline methods (DUSt3R~\cite{wang2024dust3r}, MonST3R~\cite{zhang2024monst3r}, DAS3R~\cite{xu2024das3r}, VGGT4D~\cite{hu2025vggt4d}, and Easi3R~\cite{chen2025easi3r} variants) use author-recommended configurations and serve as performance upper bounds.


\noindent\textbf{Quantitative results.}
Table~\ref{tab:davis_seg} presents the results on three DAVIS benchmarks. In the offline setting, methods like VGGT4D and Easi3R\_monst3r establish strong upper bounds by leveraging bidirectional global attention over the entire video sequence. However, performance typically degrades significantly in the streaming setting due to the absence of future context and the inherent difficulty of incremental temporal reasoning. Existing streaming methods (CUT3R, Stream3R, StreamVGGT), originally designed for geometry estimation without native segmentation support, struggle to produce accurate dynamic object masks even with our carefully adapted implementation, highlighting a fundamental limitation of purely geometric streaming architectures.

Our \textbf{Stream4D} substantially outperforms all streaming baselines across all three benchmarks, narrowing the gap with offline methods. Compared to the strongest streaming competitor StreamVGGT, Stream4D achieves absolute gains of $+8.48$ ($43.39$ vs.\ $34.91$) on DAVIS-2016, $+9.12$ ($46.45$ vs.\ $37.33$) on DAVIS-2017, and $+9.86$ ($43.06$ vs.\ $33.20$) on DAVIS-all in terms of mean IoU ($\mathcal{J}_M$). Similar improvements are consistently observed in recall ($\mathcal{J}_R$), demonstrating the effectiveness of our unified architecture for joint geometry and segmentation under strict streaming constraints.

\noindent\textbf{Qualitative results.}
Fig.~\ref{fig:qual_seg} presents representative sequences comparing the input frame, Easi3R masks, and our masks. Due to pairwise adjacent statistics only, Easi3R is prone to temporal jitter and boundary breaks under short-term occlusions and non-rigid deformations. In contrast, our streaming multi-level temporal aggregation with consistency constraints maintains stable segmentation and complete boundaries across both slow motions (e.g., camel, goat, pig, dog) and fast motions (e.g., skateboarder, motorcyclist); masks evolve smoothly over time and align with true motion patterns, demonstrating the robustness and adaptability of our causal attention-based approach in online settings. More visualized results can be found in the appendix.

\begin{figure}[t]
    \centering
    \includegraphics[width=0.9\linewidth]{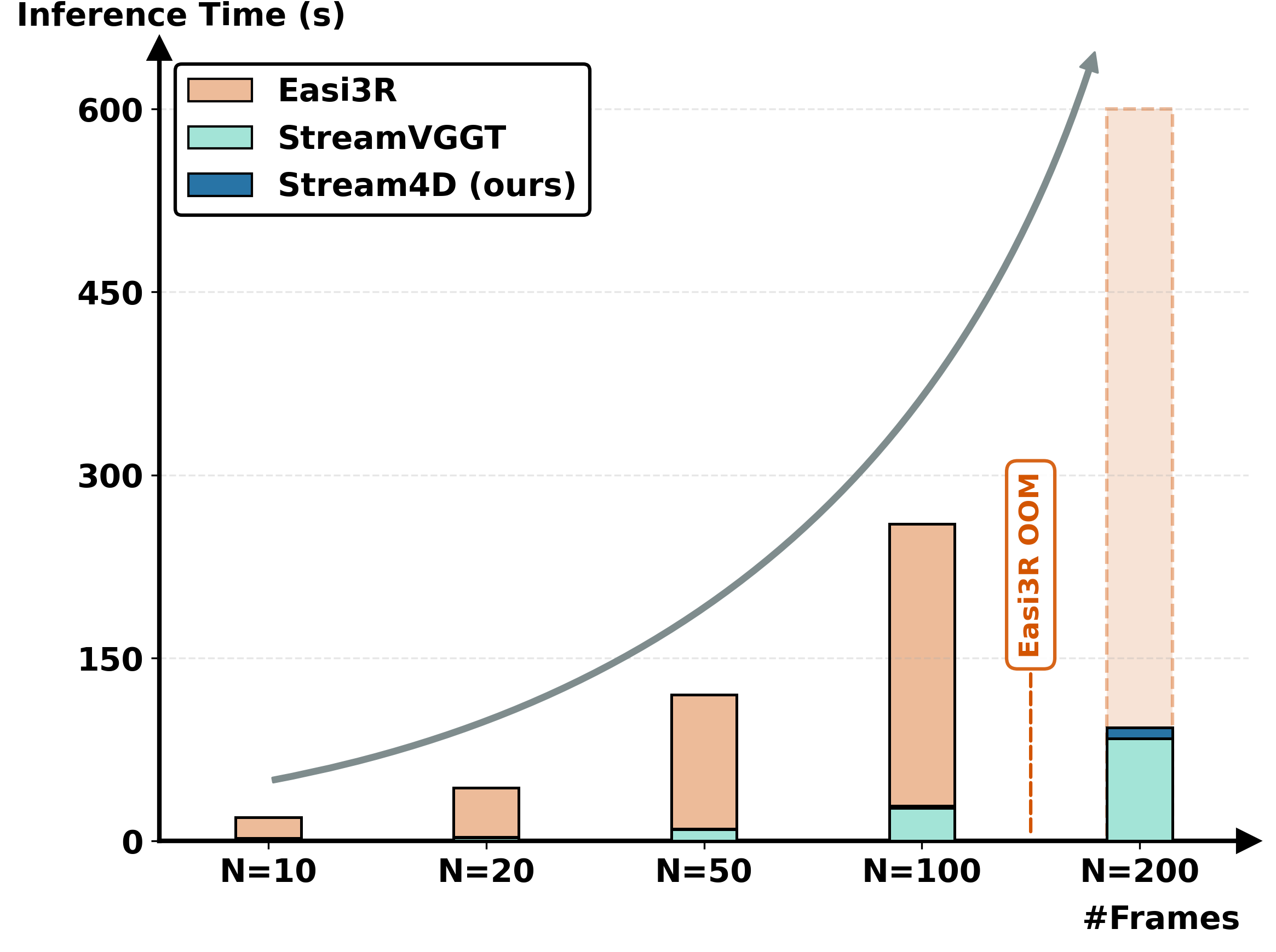}
    \caption{\textbf{Inference time comparison across different sequence lengths.} 
  While Easi3R exhibits exponential time growth and suffers from Out-of-Memory (OOM) errors between 100 and 200 frames, Stream4D maintains linear scalability, performing comparably to the efficient StreamVGGT baseline.}
    \label{fig:inference_time}
\end{figure}

\subsection{4D Reconstruction}
\label{4D}

We evaluate 4D reconstruction performance on the DyCheck dataset, a challenging benchmark comprising real-world monocular videos with complex object motion, primarily focusing on camera pose estimation accuracy. Following prior work, we report standard pose metrics including Absolute Trajectory Error (ATE), Relative Translation Error (RTE), and Relative Rotation Error (RRE). All streaming methods operate in single-pass causal mode, strictly forbidding access to future frames to ensure a fair and realistic evaluation protocol.

\noindent\textbf{Quantitative results.}
Table~\ref{tab:dycheck_recon} summarizes results on DyCheck. Offline methods such as VGGT4D achieve strong performance by leveraging full-sequence optimization, improving over their respective bases (VGGT, MonST3R, DUSt3R). Among streaming methods, CUT3R---despite being extensively trained across diverse datasets---serves as a competitive concurrent baseline. Our method (Stream4D), operating training-free and within causal streaming constraints, delivers accurate and complete reconstructions comparable to or better than most baselines, and on pose metrics achieves consistent improvements over Stream3R (ATE $0.014$, $\downarrow$6.7\%; RTE $0.017$, $\downarrow$5.6\%; RRE $0.539$, $\downarrow$1.1\%), while maintaining identical reconstruction metrics to isolate the effect of motion decoupling.

\noindent\textbf{Qualitative results.}
Fig.~\ref{fig:qual_recon_comp} compares the reconstruction quality of Stream3R~\cite{lan2025stream3r} and Stream4D on representative dynamic sequences. As shown, the baseline Stream3R exhibits severe entanglement between dynamic and static elements, leading to geometric corruption, appearance distortion, and ghosting artifacts in regions of significant motion. In contrast, our Stream4D significantly improves reconstruction fidelity in this challenging streaming setting. The improvement stems from two key designs: (1) attention-based dynamic segmentation that robustly separates motion from the static background without requiring external supervision, and (2) soft key masking that suppresses dynamic interference during historical aggregation. More visualized results are provided in the appendix.

\noindent\textbf{Ghosting suppression.}
 A core failure mode of causal streaming reconstruction in dynamic scenes is temporal over-integration: as each new frame attends to and fuses information from the past, moving objects are repeatedly written into the scene, creating stacked geometric replicas along their motion paths (ghosting artifacts). 
This artifact intensifies with longer sequences and may eventually bleed into nearby static surfaces, degrading ground, building facades, and road textures.
As illustrated in Fig.~\ref{fig:ghosting}, baseline streaming methods exhibit clear motion trails---pedestrians and vehicles appear multiple times at different locations, forming overlapping silhouettes and blurred structures.

Stream4D suppresses this accumulation by decoupling motion from the causal fusion process. 
Specifically, we aggregate temporal saliency from causal attention to derive dynamic masks on-the-fly, and apply soft key masking to suppress dynamic regions in the cached keys. This prevents dynamic features from dominating historical aggregation, so that the static background remains stable and sharp, and dynamic objects are retained only in the current frame without overlapping replicas, even over long streaming sequences.




\subsection{Efficiency and Scalability}
\label{Efficiency}

We evaluate the inference time and scalability of our proposed method across varying image sequence lengths ($N \in \{10, 20, 50, $ $100, 200\}$) with an input resolution of $512 \times 384$ on a single NVIDIA A100 GPU. Figure~\ref{fig:inference_time} illustrates the comparison between Easi3R~\cite{chen2025easi3r}, StreamVGGT~\cite{zhuo2025streaming}, and our Stream4D. As shown, Easi3R~\cite{chen2025easi3r} suffers from severe computational bottlenecks due to its reliance on global pairwise attention. Its inference time grows quadratically, reaching 260.0s at $N=100$. Crucially, due to its heavy memory footprint, Easi3R~\cite{chen2025easi3r} fails to process longer sequences entirely, triggering an \textit{Out-of-Memory} (OOM) error before reaching $N=200$.

In contrast, Stream4D demonstrates robust scalability with near-linear growth in computation. By leveraging a causal architecture similar to StreamVGGT, Stream4D maintains efficient inference even for long sequences. At $N=200$, Stream4D achieves an inference time of 93.1s, which is comparable to the highly efficient StreamVGGT baseline (84.1s). The marginal increase in latency ($\approx$10\%) is attributed to the lightweight overhead introduced by our dynamic segmentation and soft key masking mechanisms. This result confirms that Stream4D successfully avoids the OOM issues of Easi3R while maintaining high computational efficiency for long-term 4D reconstruction, making it highly practical for continuous real-world video applications.

\section{Conclusion}
We present the Stream4D, a training-free framework for motion-decoupled 4D scene reconstruction tailored to causal streaming architectures. Our key insight is that causal attention implicitly encodes rich temporal motion cues within the KV cache. Leveraging multi-level attention aggregation with depth priors, we derive reliable dynamic masks that decouple static and dynamic regions. We then apply soft key masking to directly suppress dynamic regions in the cached keys, preventing dynamic features from interfering with historical aggregation. This design improves camera pose estimation accuracy and helps mitigate ghosting artifacts over long sequences, all under strict single-pass, causal inference constraints. Experiments on DAVIS and DyCheck demonstrate that Stream4D achieves competitive or superior performance compared to existing streaming baselines in both dynamic segmentation and pose estimation, while maintaining efficient inference for long-sequence 4D reconstruction.

\noindent\textbf{Limitations and future work.}
While Stream4D demonstrates the potential of mining motion cues from causal attention, current aggregation relies primarily on attention statistics and depth priors. Future work will explore richer cue extraction from the KV cache, such as incorporating cross-scale temporal patterns and semantic-aware saliency. Beyond static-scene reconstruction, we aim to extend the framework toward dynamic object reconstruction by leveraging the decoupled motion representations.

\bibliographystyle{ACM-Reference-Format}
\bibliography{sample-base}










\end{document}